\documentclass[conference]{IEEEtran}

\usepackage{amssymb}
\usepackage{subfigure}
\usepackage{makecell}
\usepackage{booktabs}
\usepackage{graphicx}
\usepackage{multirow}
\usepackage{amsmath}
\usepackage{soul}
\usepackage{hyperref}
\usepackage{array}
\usepackage{booktabs}

\usepackage{float}
\usepackage{algorithm}
\usepackage{algorithmicx}
\usepackage{algpseudocode}

\usepackage{amsmath,amssymb,amsfonts}
\usepackage{bbm}

\usepackage{makecell}

\begin{document}

\title{Adaptive Label Correction for Robust Medical Image Segmentation with Noisy Labels}
\author{
\textbf{Chengxuan Qian}\textsuperscript{1}, 
\textbf{Kai Han}\textsuperscript{1},
\textbf{Jianxia Ding}\textsuperscript{1},
\textbf{Chongwen Lyu}\textsuperscript{1},
\textbf{Zhenlong Yuan}\textsuperscript{2},
\textbf{Jun Chen}\textsuperscript{1}, 
\textbf{Zhe Liu}\textsuperscript{1,$\dagger$}
\vspace{0.5em}
\\
{\small
\textsuperscript{1}Jiangsu University ~~~
\textsuperscript{2}University of Chinese Academy of Sciences~~~
}
\\
{\small
chengxuan.qian@stmail.ujs.edu.cn, zliu@ujs.edu.cn
}
\vspace{-1em}
}


\maketitle

\begin{abstract}

Deep learning has shown remarkable success in medical image analysis, but its reliance on large volumes of high-quality labeled data limits its applicability. While noisy labeled data are easier to obtain, directly incorporating them into training can degrade model performance. To address this challenge, we propose a Mean Teacher-based Adaptive Label Correction (ALC) self-ensemble framework for robust medical image segmentation with noisy labels. The framework leverages the Mean Teacher architecture to ensure consistent learning under noise perturbations. It includes an adaptive label refinement mechanism that dynamically captures and weights differences across multiple disturbance versions to enhance the quality of noisy labels. Additionally, a sample-level uncertainty-based label selection algorithm is introduced to prioritize high-confidence samples for network updates, mitigating the impact of noisy annotations. Consistency learning is integrated to align the predictions of the student and teacher networks, further enhancing model robustness. Extensive experiments on two public datasets demonstrate the effectiveness of the proposed framework, showing significant improvements in segmentation performance. By fully exploiting the strengths of the Mean Teacher structure, the ALC framework effectively processes noisy labels, adapts to challenging scenarios, and achieves competitive results compared to state-of-the-art methods.

\end{abstract}

\section{Introduction}

\begin{figure}[ht]
	\centering
	\includegraphics[width=0.95\columnwidth]{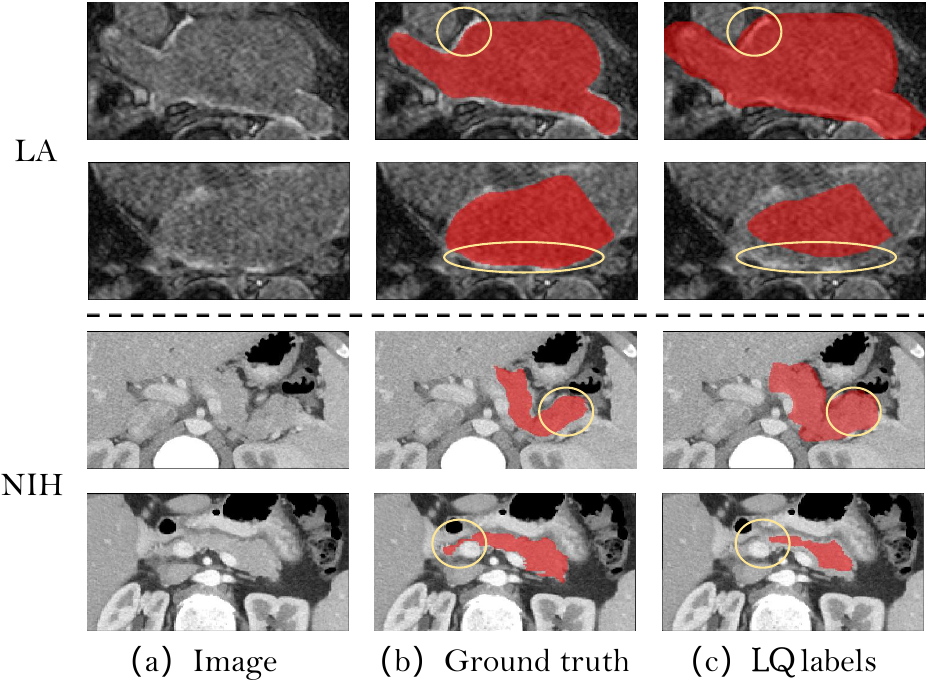}
        \vspace{-0.5em}
	\caption{Visualization of noisy labels: The first and third rows display LQ labels from ground truth expansion, while the second and fourth rows depict those from ground truth erosion.}
        \vspace{-1.5em}
	\label{fig1}
\end{figure}

Medical image segmentation is a fundamental task in computer-aided diagnosis, guiding critical processes such as disease localization, treatment planning, and surgical navigation \cite{gao2025medical, wu2025medical,jing2025multi,han2025region}. While deep learning has significantly advanced this field, its success relies heavily on large volumes of high-quality annotated data \cite{azad2024medical, hao2024ssdc, heidari2023hiformer, han2022effective,xing2025re,qian2025decalign,qian2025dyncim,chen2025haif,yuan2025dvp}. However, the performance of these deep models relies heavily on large-scale, high-quality annotated datasets, which are difficult to obtain in the medical domain due to the time-consuming, expertise-intensive nature of manual annotation and the inherent variability in imaging modalities and expert interpretations, often resulting in noisy or inconsistent labels \cite{han2024deep,dong2025uncertainty,wu2025dual,chen2025multi}. As a result, many practical scenarios rely on weak supervision, crowd-sourcing, or machine-generated pseudo labels, which are more cost-effective but often introduce label noise. Figure \ref{fig1} illustrates examples of such noisy labels, showing discrepancies caused by ground truth expansion and erosion. These labeling inaccuracies pose a significant challenge, as training models directly on noisy annotations can lead to performance degradation, overfitting to incorrect labels, and poor generalization.

To address the problem of noisy labels in medical image segmentation, recent research has explored various learning paradigms. Broadly, existing methods can be categorized into two groups: label quality-agnostic and label quality-aware approaches. Label quality-agnostic methods treat all labels equally without distinguishing between their reliability. For instance, approaches like confident learning integrated into teacher-student frameworks aim to identify corrupted labels indirectly during training \cite{zhang2020characterizing}. Similarly, dual-network schemes leveraging mutual distillation \cite{fang2023reliable} have shown promise in mitigating the adverse effects of noise. Some methods focus on weak supervision, utilizing minimal annotations (e.g., sparse points) with self-training or cross-monitoring mechanisms to refine pseudo labels progressively \cite{zhai2023pa}. In contrast, label quality-aware methods explicitly assess and utilize the quality of annotations to guide learning \cite{xu2021noisy, qian2025dyncim, dolz2021teach}. These techniques often employ uncertainty estimation (e.g., via Monte Carlo dropout or model ensembling) to quantify label confidence and apply selective training strategies. For example, anti-noise methods \cite{xu2022anti} have demonstrated the effectiveness of such awareness, often outperforming agnostic approaches by emphasizing reliable supervision. Despite these advances, existing methods still face notable limitations. Most importantly, they often rely on static or heuristic measures of uncertainty, failing to capture nuanced variations across multiple perturbation versions of the data. Furthermore, many approaches lack a self-ensemble mechanism capable of dynamically adapting to evolving label quality during training.

\begin{figure*}[t]%
	\centering
	\includegraphics[width=0.7\textwidth]{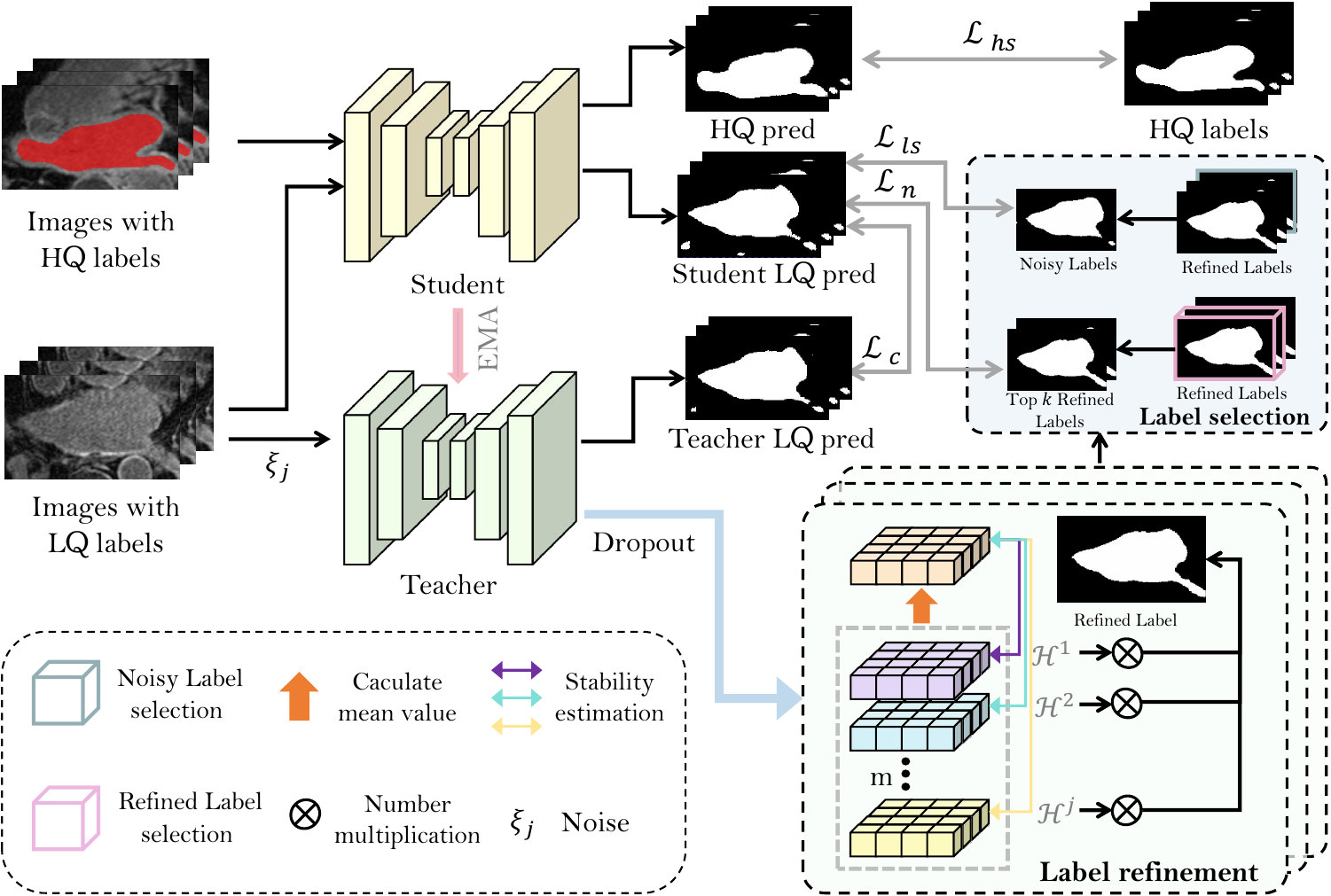}
	\caption{Overview of our proposed ALC approach, which employs a student-teacher network to integrate label correction and selection. The total loss comprising high-quality label loss $\mathcal{L}_{hs}$, low-quality label loss $\mathcal{L}_{ls}$, noisy label loss $\mathcal{L}_{n}$, and consistency loss $\mathcal{L}_{c}$. }\label{fig2}
    \vspace{-1em}
\end{figure*}

In this paper, we propose an Adaptive Label Correction (ALC) self-ensemble architecture for robust medical image segmentation with noisy labels, leveraging the Mean Teacher structure to effectively address the challenges of noisy labels. Unlike traditional methods that rely primarily on uncertainty to evaluate label quality, our approach incorporates an adaptive label correction mechanism that dynamically captures and weights differences between various perturbation versions, improving label reliability. To further strengthen the trustworthiness of the training process, we introduce a sample-level uncertainty-based label selection strategy, prioritizing high-confidence samples for gradient updates. The Mean Teacher framework, with its consistency learning and lightweight design through exponential moving average (EMA), enables the efficient extraction of latent semantic information from low-quality labels, facilitates dynamic noise correction, and ensures computational efficiency. Extensive experiments validate that our method surpasses state-of-the-art approaches in handling noisy label scenarios, providing enhanced robustness and adaptability for medical image segmentation. Specifically,

\begin{itemize}
    \item We propose ALC, a novel self-ensemble framework that integrates adaptive label correction, sample-level selection, and consistency learning to robustly handle noisy labels in medical image segmentation.
    \item By capturing perturbation-based discrepancies and leveraging dynamic sample selection, our method offers a principled approach to refine label quality and prioritize trustworthy supervision.
    \item Extensive experiments on public datasets confirm the superiority of ALC over existing state-of-the-art methods, showcasing its robustness, adaptability, and practical value for real-world medical imaging applications.
\end{itemize}

\section{Related work}\label{sec2}
\subsection{Medical image segmentation}
In recent years, advancements in deep learning have significantly progressed various medical image segmentation tasks \cite{chen2022recent}. U-Net \cite{ronneberger2015u}, a milestone network, is widely used in medical image segmentation tasks due to its simple and powerful design. Subsequently, various networks based on U-Net continued to emerge \cite{punn2022modality}, such as U-Net3+ \cite{huang2020unet}, nnU-Net \cite{isensee2018nnu}, V-Net \cite{milletari2016v}, etc. In addition, vision transformers (ViT)\cite{shamshad2023transformers} have been integrated into convolutional neural networks (CNNs) due to their excellent ability to capture global information \cite{chen2021transunet,cao2022swin,hatamizadeh2022unetr}. Recently, language modality data have been used to enhance and guide medical image segmentation due to limited image data \cite{li2023lvit,lee2023text}. Although deep learning has made significant progress in the field of medical image segmentation, the performance of these methods can be adversely affected by the presence of noisy labels, highlighting the need for effective strategies to address this challenge.

\subsection{Segmentation with noisy labels}
Leveraging noisy label data is a crucial research direction in the domain of medical image analysis. Existing methods for handling noisy labels can be broadly categorized into label selection and label correction. The essence of label selection lies in evaluating the quality of noisy labels. Assessing the quality of noisy labels allows for a better understanding of their potential impact on model training. \cite{zhu2019pick} designed a label quality evaluation strategy by assessing the conflict between noisy labels and consistency in clean labels. Based on Co-Teaching \cite{han2018co}, Tri-network \cite{zhang2020robust} introduced three networks for noisy label learning, where each of the two networks alternately selects more information data for the third network. \cite{shi2021distilling} estimated and selected noisy labels from both an image-level and pixel-level perspective. Similarly, \cite{xue2022robust} utilized a self-integrating model with noisy labeled filters to select clean and noisy samples and adopted a co-training strategy based on global-local representation learning by training with clean label samples. To select high-quality pseudo labels, \cite{yang2024non} adopted an entropy regularization to obtain high-confidence pseudo masks for efficient training. Although the above label selection methods are simple and effective, they can easily lead to the loss of information from filtered noisy labels, resulting in suboptimal segmentation results.

Label correction methods aim to algorithmically rectify the incorrect portions of noisy labels. Taking into account the intra-class and inter-class affinity relationships, \cite{guo2022joint} designed a differential affinity reasoning (DAR) module to correct pixel-level segmentation predictions by reasoning about the intra-class and inter-class affinity relationships. \cite{bu2023noisy} utilized an additional label correction network to learn the label denoising process to correct the noise in the labels. \cite{guo2023sac} combined the attention network and constraint network to deal with noisy labels. where the former could solve the problem of data-dependent denoising while the latter could alleviate the over-fitting problem caused by insufficient data. To address the problem of excessive dependence on model outputs during noisy label correction, \cite{liu2024region} introduced a Region-Scalable Fitting (RSF) framework, which integrates image features from RSF to facilitate the label correction process. To address the training challenges posed by datasets with limited or missing supervision signals, \cite{wang2024mixsegnet} designed a system comprising a Vision Transformer (ViT) and a Convolutional Neural Network (CNN), which work collaboratively through effective strategies such as network self-assembly and dynamic label assembly.

Although the above methods have achieved better performance on certain segmentation tasks, these methods may be complex and require fine parameter tuning and model selection. In comparison, our method utilizes noisy labels from two perspectives: label refinement and label selection, fully leveraging the advantages of both to effectively handle noisy label issues.

\section{Methodology}

The proposed Adaptive Label Correction (ALC) framework effectively mitigates the impact of noisy labels by leveraging HQ labels for reliable supervision and refining LQ labels to uncover latent semantic information. The framework achieves this through three core components: 1) \textbf{HQ Label Learning}, which provides a solid foundation for accurate segmentation using clean annotations; 2) \textbf{LQ Label Learning}, which refines noisy labels through adaptive weighting and selects high-confidence samples via uncertainty-based evaluation to reduce noise interference; and 3) \textbf{Consistency Learning}, which utilizes a student-teacher network to align predictions under perturbations, enhancing model robustness. By dynamically integrating these components, the framework extracts optimal supervision and semantic insights, delivering superior performance in medical image segmentation.

\subsection{Learning from high-quality labeled data}

Given a dataset $D$ comprising a high-quality (HQ) label dataset $D_{h}=\{(X^i_{h},Y^i_{h})\}_{i=1}^{N_{h}}$ and a low-quality (LQ) label dataset $D_{n}=\{(X^i_{n},Y^i_{n})\}_{i=1}^{N_{n}}$, where $X$, $Y$ represent images and their corresponding labels, and $N_{h}$ and $N_{n}$ denote the numbers of HQ and LQ label samples, receptively. Our objective is to leverage HQ labels for reliable baseline predictions and refining LQ labels to extract latent semantic information, enhancing overall model learning.


To tackle the challenges of noisy and inconsistent annotations in medical image segmentation, the proposed framework leverages high-quality (HQ) labels to provide reliable supervision and establish a robust baseline. HQ labels ensure accurate and consistent training by guiding the student network to minimize the segmentation loss:

\begin{equation}\label{eq1}
	\mathcal{L}_{hs}= \frac{1}{B_h}\sum_{i=1}^{B_h}\ell_{seg}(f_{\theta s}(X^i_{h}),Y^i_{h})
\end{equation}

where $f_{\theta s}$ represents the student network parameterized by $\theta_s$, $B_h$ is the batch size of HQ labeled data. The segmentation loss $\ell_{seg}$ is formulated as a weighted combination of Cross Entropy loss $\ell_{ce}$ and Dice loss $\ell_{dice}$:

\begin{equation}\label{eq2}
	\ell_{seg}=\frac{1}{2}(\ell_{ce}+\ell_{dice})
\end{equation}

The loss formulation combines \( \ell_{ce} \) for pixel-wise accuracy and \( \ell_{dice} \) for structural alignment, balancing fine-grained precision with segmentation robustness. This approach leverages HQ labels for consistent supervision, establishing a strong baseline to effectively integrate LQ labels and their latent semantic information.

\subsection{Learning from low-quality label data}

The proposed framework addresses the challenges of low-quality (LQ) labels by introducing two key processes: Label Refinement and Label Selection. These processes aim to refine noisy annotations and prioritize reliable samples, effectively mitigating the adverse effects of label noise while fully leveraging the latent semantic information in LQ data.

\subsubsection{Label Refinement}

Low-quality labels often suffer from noise and inconsistencies, requiring an effective refinement process to enhance their reliability. The framework employs multiple dropout operations \cite{kendall2017uncertainties} to simulate perturbations and extract stable features, leveraging the teacher network $f_{\theta t}$ to generate $m$ probabilistic outputs $\Psi$:

\begin{equation}\label{eq3}
	\Psi^{i,j}= f_{\theta t}(X^i_{n}+\xi_j), j\in \{{1,\dots,m}\}
\end{equation}

where $i$, $j$, $m$ and $\xi$ represent the sample index, dropout index, number of dropout operations, and Gaussian noise, respectively. The voxel-level mean value $\Psi_{avg}$ is then computed across all dropout outputs:

\begin{equation}\label{eq4}
	\Psi_{avg}^{i,(x,y,z)}= \frac{1}{m}\sum_{j=1}^{m}\Psi^{i,j,(x,y,z)}
\end{equation}

where $(x,y,z)$ denotes the spatial location index. To evaluate the stability of outputs at each voxel, Kullback–Leibler (KL) divergence is used:

\begin{equation}\label{eq5}
	\begin{aligned}
		\mathcal{H}^{j,(x,y,z)}&\approx KL(\Psi_{avg}^{i,(x,y,z)}\Vert \Psi^{i,j,(x,y,z)})\\&=\Psi_{avg}^{i,(x,y,z)}log \frac{\Psi_{avg}^{i,(x,y,z)}}{\Psi^{i,j,(x,y,z)}}
	\end{aligned}
\end{equation}

A higher $\mathcal{H}^{j,(x,y,z)}$ value indicates greater uncertainty for the $j$-th output at voxel $(x,y,z)$. To adaptively refine labels, the outputs are fused using weights inversely proportional to their uncertainty:

\begin{equation}\label{eq6}
	\hat{Y}_{ref}^{i,(x,y,z)}=\mathop{\arg max}\limits_{k\in\{1,2,\dots,C\}}[\frac{1}{m}\sum_{j=1}^{m}\frac{\Psi^{i,j,(x,y,z)}e^{-\mathcal{H}^{j,(x,y,z)}}}{\sum_{j}e^{-\mathcal{H}^{j,(x,y,z)}}}]
\end{equation}

where $C$ is the number of segmentation classes. This refinement process improves label quality by dynamically weighting predictions based on their reliability.

\subsubsection{Label selection}

Despite label refinement, not all samples achieve sufficient quality for effective training. To further ensure reliability, a sample-level uncertainty-based selection algorithm is introduced. The uncertainty score $\mathcal{U}^i$ for each sample is computed as:

\begin{equation}\label{eq7}
	\mathcal{U}^i=\frac{1}{N_{vox}}\sum_{x,y,z}\frac{1}{m}[\sum_{j=1}^{m}\Psi^{i,j,(x,y,z)}-(\frac{1}{m}\sum_{j=1}^{m}\Psi^{i,j,(x,y,z)})]^2
\end{equation}

Here, $N_{vox}=W \cdot H \cdot Z$ represents the total number of voxel in the sample. $W$, $H$ and $Z$ denote the spatial dimensions. Samples with the lowest uncertainty scores $\mathcal{U}^i$ are prioritized for training. The top $k$ samples are selected to compute the low-quality label loss $\mathcal{L}_{ls}$:

\begin{equation}\label{eq8}
	\mathcal{L}_{ls} = \frac{1}{k}\sum_{i=1}^{k}\ell_{seg}(f_{\theta s}(X^i_{n}),\hat{Y}_{ref})
\end{equation}

where $\ell_{seg}$ represents the segmentation loss. $B_n$ is the batch size of noisy label data. Remaining samples, which may still contain noise, are used for noisy label loss $\mathcal{L}_n$:

\begin{equation}\label{eq9}
	\mathcal{L}_{n} = \frac{1}{B_n-k+1}\sum_{i=k+1}^{B_n}\ell_{seg}(f_{\theta s}(X^i_{n}),\hat{Y}_{n})
\end{equation}



\subsection{Consistency learning}

To enhance model robustness and mitigate the effects of noise perturbations, the framework employs consistency learning through a student-teacher network architecture. This mechanism encourages alignment between the outputs of the student $f_{\theta_s}$ and teacher $f_{\theta_t}$ networks under different perturbation conditions. The consistency loss is defined as:

\begin{equation}\label{eq10}
	\mathcal{L}_{c} = \frac{1}{N_{vox}}\sum_{x,y,z}\Vert \sum_{j=1}^{m}f_{\theta t}(X_{n}^{(x,y,z)})- f_{\theta s}(X_{n}^{(x,y,z)}) \Vert
\end{equation}

It ensures that predictions from both networks remain stable and consistent across multiple dropout operations. Here, $N_{vox}$ denotes the total number of voxels and $m$ denotes the number of perturbation iterations.


\renewcommand{\algorithmicrequire}{\textbf{Input:}}  
\renewcommand{\algorithmicensure}{\textbf{Output:}} 
\begin{algorithm}[ht]
	\caption{The training process of the overall framework.} %
	\begin{algorithmic}[1]
		\Require
		HQ label dataset $(X_h,Y_h) \in D_h$, LQ label dataset $(X_n,Y_n) \in D_n$, $k$, model parameters $\theta_t$ and $\theta_s$.
		\Ensure
		model parameters $\theta_s$ for inference.
		\State t $\gets$ 0
		\Repeat
		\State $t\gets t+1$;
		\For{sample $(X_i,Y_i)$ from $D_h$} \Comment{HQ label learning}
		\State Compute $\mathcal{L}_{hs}$ by Eq.(1); 
		\EndFor
		\For{sample $(X^n,Y^n)$ from $D_n$}; \Comment{LQ label learning}
		\State Evaluate voxel uncertainty by Eq. (3)-(5);
		
		\State Refine label by Eq. (6);
		\State Compute sample stability $\mathcal{U}$ by Eq. (7);
		\State Select top $k$ samples based on $\mathcal{U}$;
		\If{sample $i \in top\ k$}
		\State Compute $\mathcal{L}_{ls}$ by Eq. (8); 
		\Else
		\State Compute $\mathcal{L}_{n}$ by Eq. (9);
		\EndIf
		\EndFor	
		\State Compute $\mathcal{L}_{c}$ by Eq. (10);\Comment{Consistency learning}
		\State Train with total loss $\mathcal{L}_{total}$ by Eq. (11);
		\Until{the model converge}.
	\end{algorithmic}\label{algor1}
\end{algorithm}

\subsection{Total loss}

The total loss integrates four key components to balance the contributions of different learning objectives: HQ label loss \( \mathcal{L}_{hs} \), LQ label loss \( \mathcal{L}_{ls} \), noisy label loss \( \mathcal{L}_{n} \), and consistency loss \( \mathcal{L}_{c} \). The total loss is expressed as:

\begin{equation}\label{eq11}
	\mathcal{L}_{total} = \mathcal{L}_{hs}+\lambda(t)(\alpha\mathcal{L}_{ls}+\beta\mathcal{L}_{n}+\mathcal{L}_{c})
\end{equation}

where \( \lambda(t) \) dynamically adjusts the weights during training, transitioning focus from HQ labels to LQ labels and consistency learning. \( \alpha \) and \( \beta \) controls the relative importance of \( \mathcal{L}_{ls} \) and \( \mathcal{L}_{n} \), their sensitivity will be further analyzed in Section \ref{ablation} and Fig. \ref{fig6}. This formulation ensures reliable supervision from HQ labels, effective utilization of LQ labels through refinement and robust learning, and stability via consistency constraints between student and teacher networks, enabling effective segmentation despite noisy annotations. The overall training process is shown in Algorithm \ref{algor1}.

\begin{table*}
    \centering
    \caption{Segmentation results on the NIH pancreas segmentation dataset compared with state-of-the-art methods.}\label{tab1}%
    \setlength{\aboverulesep}{0pt} 
    \setlength{\belowrulesep}{0pt} 
    \setlength{\extrarowheight}{0.5pt} 
    \resizebox{\linewidth}{!}{\begin{tabular}{@{}c|c|c|c|c||c|c|c|c@{}}
            \Xhline{1.5pt}
            \multirow{2}*{Methods}  & \multicolumn{4}{c||}{10\% Set-HQ and 90\% Set-LQ} & \multicolumn{4}{c}{20\% Set-HQ and 80\% Set-LQ} \\ 
            \cline{2-9}
             & Dice [\%]$\uparrow$ & Jaccard [\%]$\uparrow$ & 95HD [mm]$\downarrow$ & ASD [mm]$\downarrow$ & Dice [\%]$\uparrow$ & Jaccard [\%]$\uparrow$ & 95HD [mm]$\downarrow$ & ASD [mm]$\downarrow$ \\[1pt]
            \hline		
            H-Sup    & 70.86 & 56.50 & 21.15 & 8.87 & 73.64 & 59.98 & 12.18 & 4.65 \\		
            HL-Sup   & 69.45 & 54.68 & 22.18 & 9.75 & 72.90 & 58.68 & 13.72 & 4.83 \\	
            \hline
            TriNet \cite{zhang2020robust}     & 73.76 & 59.59 & 21.48 & 6.77 & 76.13 & 62.23 & 9.35 & 3.84 \\
            2SRnT \cite{zhang2020characterizing}     & 72.56 & 58.13 & 20.40 & 6.08 & 75.72 & 61.66 & 11.23 & 4.32 \\
            PNL \cite{zhu2019pick}   & 72.67 & 57.93 & 19.42 & 6.19 & 76.95 & 63.22 & 9.57 & 3.83 \\
            MS-TFAL \cite{cui2023rectifying}  & 72.83 & 58.63 & 18.43 & 6.24 & 75.56 & 61.36 & 10.05 & 4.15 \\
            TBraTS \cite{zou2022tbrats}     & 71.19 & 56.92 & 23.06 & 7.06 & 75.28 & 61.21 & 8.69 & 4.95 \\
            \hline
            Decoupled \cite{luo2020semi}   & 73.16 & 59.21 & 19.79 & 5.98 & 78.13 & 64.90 & 9.33 & 3.48 \\
            KDEM \cite{dolz2021teach}  & 73.57 & 59.57 & 18.89 & 5.77 & 77.36 & 64.01 & \underline{6.24} & 2.73 \\
            MTCL-Hard \cite{xu2022anti}   & \underline{74.88} & \underline{60.49} & 19.50 &  \underline{5.54} & \underline{78.97} & 65.86 & 9.24 & \underline{2.13} \\
            MTCL-FS \cite{xu2022anti}  & 73.95 & 60.01 & 18.33 & 6.17 & 77.65 & 64.17 & 9.83 & 3.07 \\
            MTCL-UDS \cite{xu2022anti} & 74.30 & 59.94 & 19.54 & 6.09 & 78.19 & \underline{65.88} & 7.21 & 2.32 \\
            MixSegNet \cite{wang2024mixsegnet}  & 74.06 & 59.66 & \underline{18.31} & 6.46 & 77.07 & 63.94 & 8.22 & 2.76 \\
            \hline
            ALC (ours)   & \textbf{75.21} & \textbf{60.94} & \textbf{18.13} & \textbf{5.37} & \textbf{79.62} & \textbf{66.32} & \textbf{5.86} & \textbf{2.02} \\
            \Xhline{1.5pt}
    \end{tabular}}
\end{table*}

\begin{table*}
    \centering
    \caption{Segmentation results on the MSD and GWS datasets compared with state-of-the-art methods.}\label{tab2}%
    \setlength{\aboverulesep}{0pt}
    \setlength{\belowrulesep}{0pt}
    \setlength{\extrarowheight}{0.5pt}
    \resizebox{\linewidth}{!}{\begin{tabular}{@{}c|c|c|c|c||c|c|c|c@{}}
            \Xhline{1.5pt}
            \multirow{2}*{Methods}  & \multicolumn{4}{c||}{MSD Dataset} & \multicolumn{4}{c}{GWS Dataset} \\ 
            \cline{2-9}
             & Dice [\%]$\uparrow$ & Jaccard [\%]$\uparrow$ & 95HD [mm]$\downarrow$ & ASD [mm]$\downarrow$ & Dice [\%]$\uparrow$ & Jaccard [\%]$\uparrow$ & 95HD [mm]$\downarrow$ & ASD [mm]$\downarrow$ \\[1pt]
            \hline		
            TriNet \cite{zhang2020robust}     & 33.94 & 25.87 & 38.43 & 10.58 & 71.37 & 55.64 & 5.27 & 1.22 \\		
            2SRnt \cite{zhang2020characterizing}     & 34.38 & 26.02 & 33.81 & 10.94 & 71.99 & 53.38 & 4.65 & 1.12 \\	
            ADELE \cite{liu2022adaptive}   & 35.13 & 27.20 & 30.26 & 8.47 & 72.53 & 56.73 & 2.88 & 0.79 \\		
            SAC-Net \cite{guo2023sac}  & 35.35 & 26.81 & 31.42 & 6.95 & 72.32 & 56.12 & 2.37 & 0.86 \\		
            BoxTeacher \cite{cheng2023boxteacher}     & 34.91 & 26.73 & 35.56 & 7.43 & 71.46 & 55.72 & 4.24 & 1.08 \\		
            MedSAM \cite{ma2024segment}     & 32.66 & 24.40 & 29.25 & 9.36 & 69.25 & 53.70 & 4.85 & 1.17 \\		
            \hline
            Decoupled \cite{luo2020semi}   & 36.28 & 27.36 & 28.68 & 8.31 & 70.08 & 52.12 & 3.37 & 0.86 \\		
            KDEM \cite{dolz2021teach}  & 37.26 & 27.92 & 25.10 & 7.51 & 72.14 & 56.57 & 3.20 & 1.01 \\		
            MTCL-Hard \cite{xu2022anti}   & 38.23 & 29.18 & 26.02 & 6.40 & 73.06 & 57.70 & 2.69 & 1.27 \\		
            MTCL-FS \cite{xu2022anti}  & 37.72 & 28.70 & 24.69 & 7.78 & 72.21 & 56.65 & 5.69 & 0.69 \\		
            MTCL-UDS \cite{xu2022anti} & 38.26 & 28.83 & 24.08 & 7.48 & 73.18 & 57.86 & 4.31 & 1.05 \\		
            MixSegNet \cite{wang2024mixsegnet}  & 38.35 & 28.86 & 27.55 & 6.54 & 72.95 & 56.86 & 2.68 & 0.75 \\		
            \hline
            Ours   & \textbf{40.38} & \textbf{30.68} & \textbf{21.85} & \textbf{6.44} & \textbf{74.98} & \textbf{59.86} & \textbf{2.35} & \textbf{0.66} \\		
            \Xhline{1.5pt}
    \end{tabular}}
\end{table*}

\section{Experiments}
\subsection{Datasets}\label{sec4.1}

\textbf{LA segmentation dataset.} The Left Atrium (LA) dataset \cite{xiong2021global} includes 100 GE-MRIs ($0.625^3$ $mm^3$ resolution). Following \cite{adiga2024anatomically}, 80 scans are for training and 20 for testing. Patches are $112 \times 112 \times 80$, with sliding window prediction for testing.

\textbf{NIH dataset.} The NIH pancreas dataset \cite{roth2015deeporgan} has 82 CT scans. We use 62 for training and 20 for testing. Intensities are rescaled to $[0, 1]$ within [-125, 275] HU, with $96^3$ patches.

\textbf{MSD Dataset.} The Medical Segmentation Decathlon (MSD) dataset consists of 281 portal venous phase CT scans annotated with tumor masks. Each slice has a resolution of 512$\times$512 pixels, with slice counts per scan ranging from 37 to 751. We randomly partitioned the dataset into 236 scans for training and 45 for testing.

\textbf{GWS Dataset.} The Gastric Wall Segmentation (GWS) dataset is a real-world noisy label dataset comprising CT plain scans from 214 gastric cancer patients collected at Zhenjiang First People's Hospital. Each image has a resolution of 512$\times$512 pixels. Notably, only 100 cases include high-quality voxel-level annotations (HQ dataset), generated by three expert radiologists.

\begin{figure}[ht]%
	\centering
	\includegraphics[width=\columnwidth]{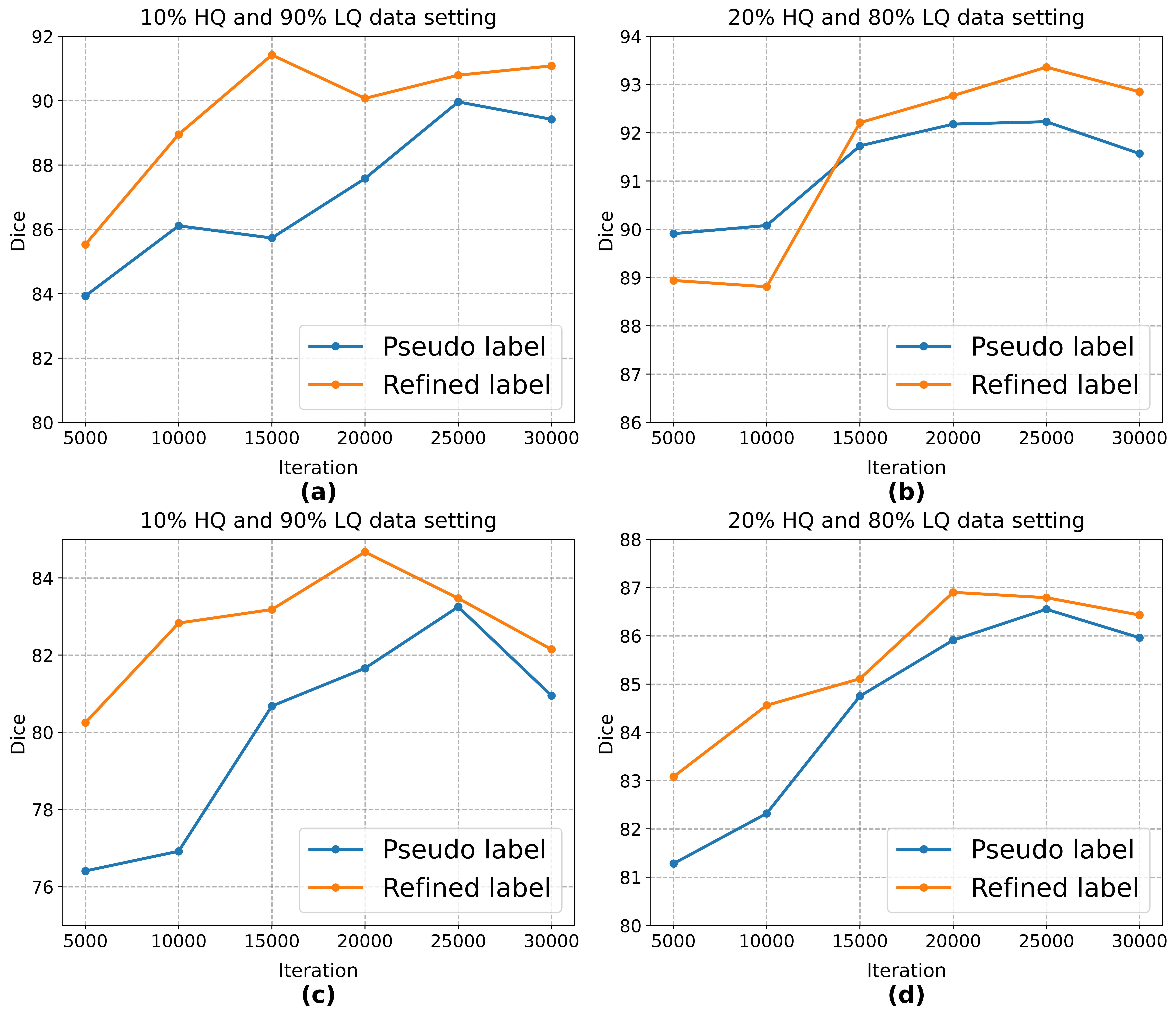}
        \vspace{-2em}
	\caption{Quantitative comparisons of pseudo and corrected label quality across iterations under varying HQ and LQ label ratios on the LA dataset (a)-(b) and NIH dataset (c)-(d).}
	\label{fig3}
\end{figure}

\subsection{Implementation Details}
To simulate annotation errors commonly made by humans, we synthetically generated noisy labels by randomly expanding or eroding the ground truth boundaries by 3–15 pixels, following the procedures outlined in \cite{xu2022anti,liu2024region}. This process introduces spatial distortions that mimic typical variations observed in manual segmentation, enabling a controlled evaluation of model robustness under label noise. For quantitative performance assessment, we adopted four widely used evaluation metrics: Dice Similarity Coefficient (Dice), Jaccard Index (Jaccard), 95\% Hausdorff Distance (95HD), and Average Surface Distance (ASD). Dice and Jaccard measure the overlap between predicted and ground truth regions, with higher values indicating better segmentation. In contrast, 95HD and ASD quantify boundary accuracy, where lower values denote superior performance. To clarify, upward arrows ($\uparrow$) denote metrics where higher is better, and downward arrows ($\downarrow$) indicate that lower values are preferred. The framework was implemented in PyTorch and run on an NVIDIA A6000 GPU with 48 GB memory. V-Net \cite{milletari2016v} served as the backbone within the Mean Teacher framework \cite{tarvainen2017mean}, with an EMA decay $\gamma$ of 0.99. Network parameters were optimized using SGD. Analysis of the $k$ hyperparameter is presented in Section \ref{sec4.5}.

\subsection{Comparison with the State-of-the-arts}


\begin{table}[ht]
	\centering
	\caption{Ablation study on LA dataset.}\label{tab3}%
	\resizebox{\linewidth}{!}{%
	\begin{tabular}{@{}c|c|c|c|c@{}}
			\Xhline{1.5pt}
			\multicolumn{5}{c}{10\% Set-HQ and 90\% Set-LQ} \\ 
			\Xhline{1pt}
			Methods & Dice [\%]$\uparrow$ & Jaccard [\%]$\uparrow$ & 95HD [mm]$\downarrow$ & ASD [mm]$\downarrow$ \\[1pt]
			\hline
			Mean-Teacher \cite{tarvainen2017mean}  & 83.79 & 72.62 & 20.07 & 4.01 \\			
			ALC w/o LS                  & 84.42 & 73.59 & 10.49 & 2.87 \\
            ALC w/o LR             & \underline{87.31} & \underline{77.81} & \underline{8.79}  & \underline{2.81} \\

			ALC (ours)                  & \textbf{88.87} & \textbf{80.23} & \textbf{8.36} & \textbf{2.29} \\
			\Xhline{1pt}
	\end{tabular}}

	\vspace{8pt} 

	\resizebox{\linewidth}{!}{%
	\begin{tabular}{@{}c|c|c|c|c@{}}
			\Xhline{1.5pt}
			\multicolumn{5}{c}{20\% Set-HQ and 80\% Set-LQ} \\ 
			\Xhline{1pt}
			Methods & Dice [\%]$\uparrow$ & Jaccard [\%]$\uparrow$ & 95HD [mm]$\downarrow$ & ASD [mm]$\downarrow$ \\[1pt]
			\hline
			Mean-Teacher \cite{tarvainen2017mean}  & 87.51 & 77.92 & 8.83  & 2.86 \\			
			ALC w/o LS                  & 88.11 & 79.19 & \underline{6.76}  & \underline{1.99} \\
ALC w/o LR                  & \underline{88.53} & \underline{79.53} & 9.13  & 2.77 \\

			ALC (ours)                  & \textbf{89.29} & \textbf{80.99} & \textbf{6.30} & \textbf{1.22} \\
			\Xhline{1pt}
	\end{tabular}}

    	\vspace{8pt} 

\resizebox{\linewidth}{!}{%
\begin{tabular}{@{}c|c|c|c|c@{}}
        \Xhline{1.5pt}
        \multicolumn{5}{c}{30\% Set-HQ and 70\% Set-LQ} \\ 
        \Xhline{1pt}
        Methods & Dice [\%]$\uparrow$ & Jaccard [\%]$\uparrow$ & 95HD [mm]$\downarrow$ & ASD [mm]$\downarrow$ \\[1pt]
        \hline
        Mean-Teacher \cite{tarvainen2017mean}  & 89.53 & 81.11 & 7.83  & 2.59 \\			
        ALC w/o LS                  & 89.68 & 81.24 & \underline{6.60}  & \underline{1.79} \\
        ALC w/o LR                  & \underline{89.83} & \underline{81.61} & 7.38  & 1.98 \\
        ALC (ours)                  & \textbf{90.16} & \textbf{82.14} & \textbf{5.44} & \textbf{1.59} \\
        \Xhline{1pt}
\end{tabular}}

\end{table}

\begin{table}[ht]
	\centering
	\caption{Ablation study on NIH dataset.}\label{tab4}%
	\resizebox{\linewidth}{!}{%
	\begin{tabular}{@{}c|c|c|c|c@{}}
			\Xhline{1.5pt}
			\multicolumn{5}{c}{10\% Set-HQ and 90\% Set-LQ} \\ 
			\Xhline{1pt}
			Methods & Dice [\%]$\uparrow$ & Jaccard [\%]$\uparrow$ & 95HD [mm]$\downarrow$ & ASD [mm]$\downarrow$ \\[1pt]
			\hline
			Mean-Teacher \cite{tarvainen2017mean}   & 70.59 & 55.48 & 23.58 & 7.43 \\			
			ALC w/o LS                  & 71.17 & 56.12 & 22.56 & 6.38 \\
			ALC w/o LR                  & \underline{73.23} & \underline{58.66} & \underline{19.67} & \underline{6.51} \\
			ALC (ours)                  & \textbf{75.21} & \textbf{60.94} & \textbf{18.13} & \textbf{5.37} \\
			\Xhline{1pt}
	\end{tabular}}

	\vspace{8pt} 

	\resizebox{\linewidth}{!}{%
	\begin{tabular}{@{}c|c|c|c|c@{}}
			\Xhline{1.5pt}
			\multicolumn{5}{c}{20\% Set-HQ and 80\% Set-LQ} \\ 
			\Xhline{1pt}
			Methods & Dice [\%]$\uparrow$ & Jaccard [\%]$\uparrow$ & 95HD [mm]$\downarrow$ & ASD [mm]$\downarrow$ \\[1pt]
			\hline
			Mean-Teacher \cite{tarvainen2017mean}   & 74.77 & 60.51 & 15.09 & 4.78 \\			
			ALC w/o LS                  & 75.48 & 61.50 & 11.07 & 3.96 \\
			ALC w/o LR                  & \underline{76.23} & \underline{62.29}& \underline{10.45} & \underline{3.74} \\
			ALC (ours)                  & \textbf{79.62} & \textbf{66.32} & \textbf{5.86} & \textbf{2.02} \\
			\Xhline{1pt}
	\end{tabular}}

    	\vspace{8pt} 

\resizebox{\linewidth}{!}{%
\begin{tabular}{@{}c|c|c|c|c@{}}
        \Xhline{1.5pt}
        \multicolumn{5}{c}{30\% Set-HQ and 70\% Set-LQ} \\ 
        \Xhline{1pt}
        Methods & Dice [\%]$\uparrow$ & Jaccard [\%]$\uparrow$ & 95HD [mm]$\downarrow$ & ASD [mm]$\downarrow$ \\[1pt]
        \hline
        Mean-Teacher \cite{tarvainen2017mean}   & 78.53 & 65.23 & 8.65 & 2.99 \\			
        ALC w/o LS                  & 79.75 & 66.79 & \underline{6.69} & \underline{2.17} \\
        ALC w/o LR                  & \underline{80.22} & \underline{67.34} & 10.04 & 2.05 \\
        ALC (ours)                  & \textbf{80.51} & \textbf{67.78} & \textbf{5.57} & \textbf{1.91} \\
        \Xhline{1pt}
\end{tabular}}

\end{table}

To comprehensively evaluate the effectiveness of our proposed ALC framework, we conducted extensive comparisons against both baseline models and a wide range of state-of-the-art methods on multiple datasets, including NIH pancreas segmentation, MSD, and GWS. These comparisons were performed under varying proportions of high-quality (HQ) and low-quality (LQ) labels to rigorously assess robustness under noisy label conditions. We first compare ALC with two baseline settings implemented using V-Net: H-Sup, which utilizes only HQ labels for training, and HL-Sup, which jointly uses HQ and LQ labels without differentiating their quality. As shown in Table~\ref{tab1} and Table~\ref{tab2}, HL-Sup underperforms compared to H-Sup, indicating that naively mixing noisy labels can degrade performance. In contrast, ALC surpasses both baselines across all metrics, estab2pecially under low HQ ratios (e.g., 10\% HQ and 90\% LQ), demonstrating its ability to effectively leverage noisy data through adaptive correction and selection.

\noindent \textbf{Comparison with Label Quality-Agnostic Methods.} We further compare ALC with several state-of-the-art label quality-agnostic methods, including TriNet \cite{zhang2020robust}, 2SRnT \cite{zhang2020characterizing}, PNL \cite{zhu2019pick}, MS-TFAL \cite{cui2023rectifying}, and TBraTS \cite{zou2022tbrats}. These methods do not distinguish between HQ and LQ labels during training. As shown in Table~\ref{tab1} and Table~\ref{tab2}, although some of these models achieve competitive performance in terms of Dice score (e.g., PNL and TriNet), they tend to suffer from higher boundary errors (95HD and ASD), particularly under high noise conditions. This indicates their limited capability in handling annotation noise robustly. By contrast, ALC achieves higher Dice and Jaccard scores, along with significantly reduced 95HD and ASD values, highlighting superior segmentation accuracy and boundary precision.

\noindent \textbf{Comparison with Label Quality-Aware Methods.} Next, we evaluate ALC against label quality-aware methods, which are specifically designed to mitigate the impact of noisy labels. These include Decoupled \cite{luo2020semi}, KDEM \cite{dolz2021teach}, and three variants of MTCL \cite{xu2022anti}: MTCL-Hard, MTCL-FS, and MTCL-UDS. Additionally, we consider the recently proposed MixSegNet \cite{wang2024mixsegnet}. As shown in the tables for both NIH and MSD datasets, these methods generally outperform agnostic models due to their capacity to model label confidence or uncertainty. However, ALC still surpasses them in almost all metrics. For instance, in the NIH dataset under 20\% HQ, ALC achieves a Dice score of 79.62\% and a Jaccard score of 66.32\%, outperforming the best competitor MTCL-Hard by 0.65\% and 0.46\%, respectively. Notably, ALC also achieves the lowest boundary error with a 95HD of 5.86 mm and ASD of 2.02 mm, indicating its effectiveness in refining noisy labels and promoting accurate boundary delineation.

\noindent \textbf{Results on MSD and GWS Datasets.} A similar performance trend is observed on the MSD and GWS datasets (see Table~\ref{tab2}). On the MSD dataset, ALC achieves the highest Dice score (40.38\%) and lowest 95HD (21.85 mm), outperforming the best previous method MixSegNet (Dice 38.35\%, 95HD 27.55 mm). On the GWS dataset, which represents real-world noisy labels, ALC again demonstrates its robustness, achieving a Dice score of 74.98\% and ASD of 0.66 mm, both the best among all methods. These results confirm the generalizability and effectiveness of ALC across synthetic and real-world noisy label settings.

\begin{figure}%
	\centering
	\includegraphics[width=\columnwidth]{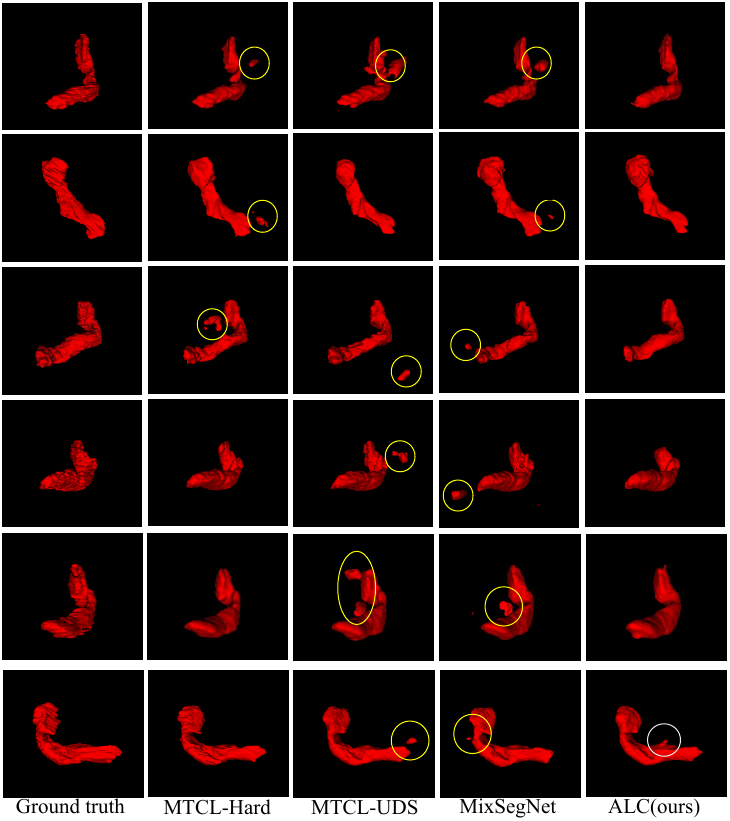} 
        \vspace{-2em}
	\caption{Visual comparison of segmentation results on the NIH dataset under the 10\% HQ and 90\% LQ label setting across different methods}\label{fig4}
    \vspace{-2em}
\end{figure}

\noindent \textbf{Visualization Analysis.} Fig.~\ref{fig4} presents a visual comparison of segmentation results from four leading methods—MTCL-Hard, MTCL-UDS, MixSegNet, and our proposed ALC—on the NIH dataset under the 10\% HQ and 90\% LQ label setting. Across all test cases, ALC consistently produces segmentation masks that most closely resemble the ground truth, accurately delineating anatomical boundaries even in challenging regions with low contrast or complex shapes. In contrast, the other methods exhibit common failure patterns, such as missing regions, boundary artifacts, or fragmented structures, as highlighted by yellow circles. These errors reflect the difficulty of learning under noisy supervision without effective label correction. ALC’s superior results are attributed to its adaptive label refinement and confidence-based sample selection, which together yield more reliable training signals. Nevertheless, some minor segmentation inaccuracies remain in ALC’s outputs, especially along intricate boundaries (indicated by white circles), suggesting occasional sensitivity of the noise-specific loss term $\mathcal{L}_n$ to extremely poor-quality labels. While these issues are limited, they highlight potential for further improving ALC’s robustness through enhanced loss formulations. Overall, the qualitative results confirm that ALC delivers not only higher segmentation accuracy but also more stable and coherent predictions in noisy label scenarios.

\subsection{Ablation Studies}\label{ablation}


To rigorously assess the individual contributions of each component in our proposed ALC framework, we conducted comprehensive ablation experiments on two benchmark datasets: the LA and NIH datasets. These experiments were designed to simulate real-world scenarios where the proportion of high-quality (HQ) labeled data is limited. Specifically, we evaluated performance under three different HQ-to-LQ label ratios: 10\%, 20\%, and 30\% HQ labels. This setup allows us to analyze how each component performs under increasing data quality and quantity. We used the Mean Teacher (MT) framework \cite{tarvainen2017mean} as a strong baseline for comparison, owing to its widely recognized effectiveness in semi-supervised learning. The experiments included: 1) \textbf{ALC w/o LS}: the framework without label selection; and 2) \textbf{ALC w/o LR}: the framework without label refinement.

As detailed in Tables \ref{tab3} and \ref{tab4}, both label selection and label refinement significantly improve segmentation performance compared to the Mean Teacher baseline. Notably, removing the label selection module (ALC w/o LS) leads to a more substantial performance degradation than removing label refinement, especially in boundary-sensitive metrics such as 95HD and ASD. For example, on the NIH dataset with 10\% HQ labels, the Dice score drops from 75.21\% (full ALC) to 71.17\% when LS is removed, while removal of LR results in a smaller decrease to 73.23\%. This pattern is consistent across both datasets and all HQ ratios. This observation highlights that sample selection based on uncertainty plays a more dominant role in mitigating the effect of noisy labels, particularly in early training stages where misguidance from LQ samples can severely impact model convergence. On the other hand, label refinement contributes to fine-grained correction of noisy labels, leading to improved boundary precision and consistency, as reflected in lower 95HD and ASD scores.

\noindent \textbf{Performance Scaling with HQ Ratio.}
Across all configurations, we observe that as the proportion of HQ data increases, overall performance improves, but the relative performance gain from ALC decreases. This trend is intuitive: with more HQ data, the reliance on LQ label correction diminishes. For instance, the Dice score of ALC on the LA dataset improves from 88.87\% (10\% HQ) to 90.16\% (30\% HQ), yet the margin of improvement over the Mean Teacher baseline narrows. This suggests that ALC is particularly beneficial in low-HQ data regimes, making it well-suited for real-world clinical scenarios where HQ annotations are limited.

\noindent \textbf{Effectiveness of Label Refinement During Training.}
Fig.~\ref{fig3} provides further evidence of the effectiveness of our ALC framework by comparing the quality of pseudo labels and refined labels throughout training. In all HQ/LQ configurations and across both datasets, refined labels consistently achieve higher Dice scores than their unrefined counterparts. This demonstrates that ALC’s adaptive label refinement mechanism substantially enhances the quality of supervision, especially in early and mid-training stages when pseudo labels are inherently noisy. The performance gap is particularly notable under the 10\% HQ setting on the NIH dataset, where refined labels outperform pseudo labels by a large margin, affirming ALC’s robustness in challenging conditions. These findings reinforce that the effectiveness of ALC stems not only from final performance metrics but also from its ability to generate more accurate and stable training signals over time, enabling it to consistently outperform state-of-the-art methods under noisy label scenarios.

In summary, our ablation studies demonstrate that both label selection and label refinement are essential and complementary components of the ALC framework. Label selection ensures that the model learns from the most reliable data at each stage, while label refinement corrects noisy annotations, leading to more accurate supervision. Their combination yields the best segmentation performance across all settings. Furthermore, ALC shows particular strength in low-HQ scenarios, highlighting its practical utility in medical image segmentation tasks where annotation quality and quantity are often constrained. Future extensions may explore adaptive weighting strategies between selection and refinement to further enhance model robustness.

\begin{figure}[ht]%
	\centering
	\includegraphics[width=\columnwidth]{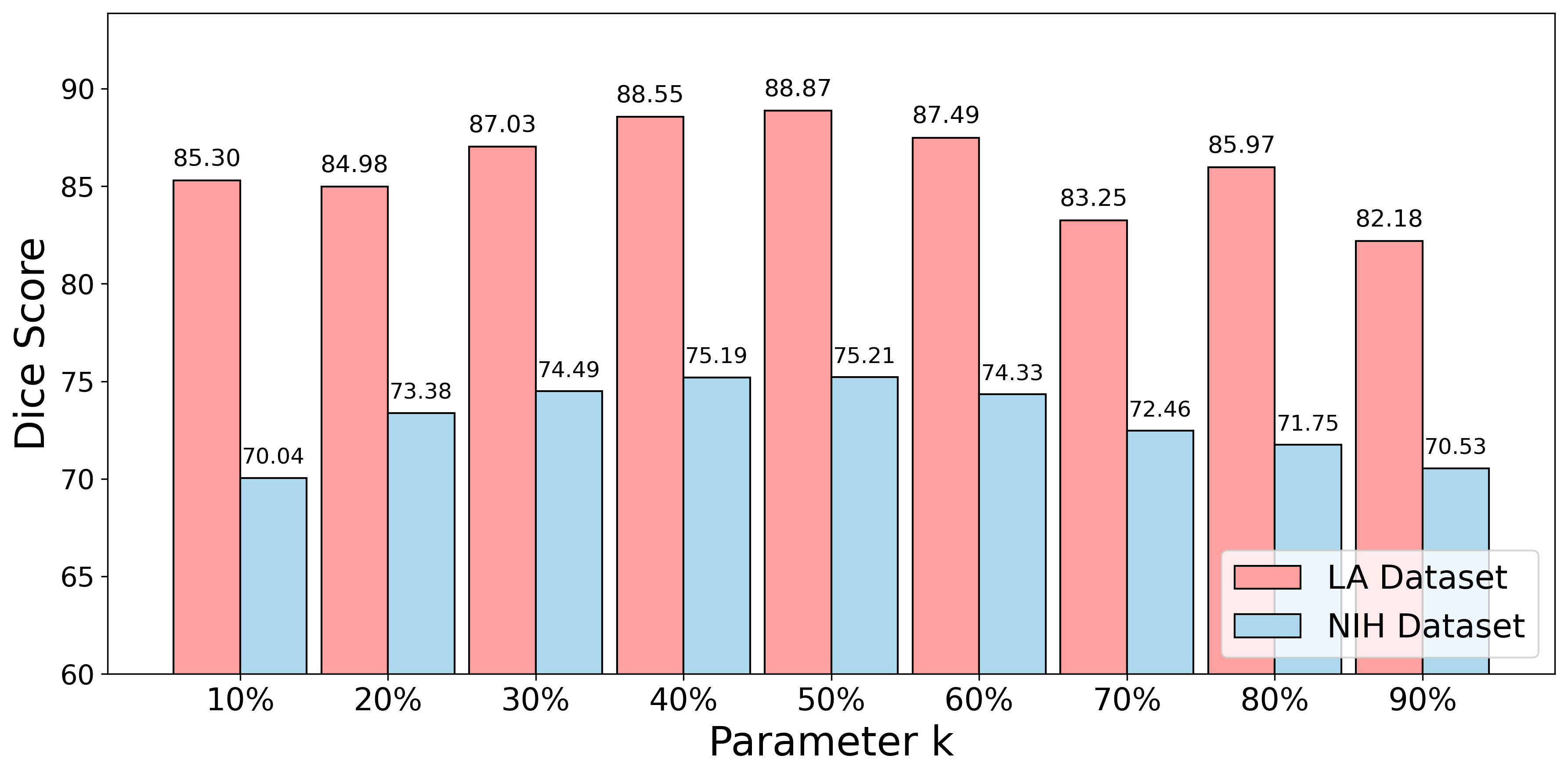}
        \vspace{-2em}
	\caption{The impact of different $k$ values on the LA and NIH datasets under 10\% HQ label data and  90\% LQ label data.}\label{fig5}
\end{figure} 

\vspace{-1em}

\begin{figure}[ht]%
	\centering
	\includegraphics[width=\columnwidth]{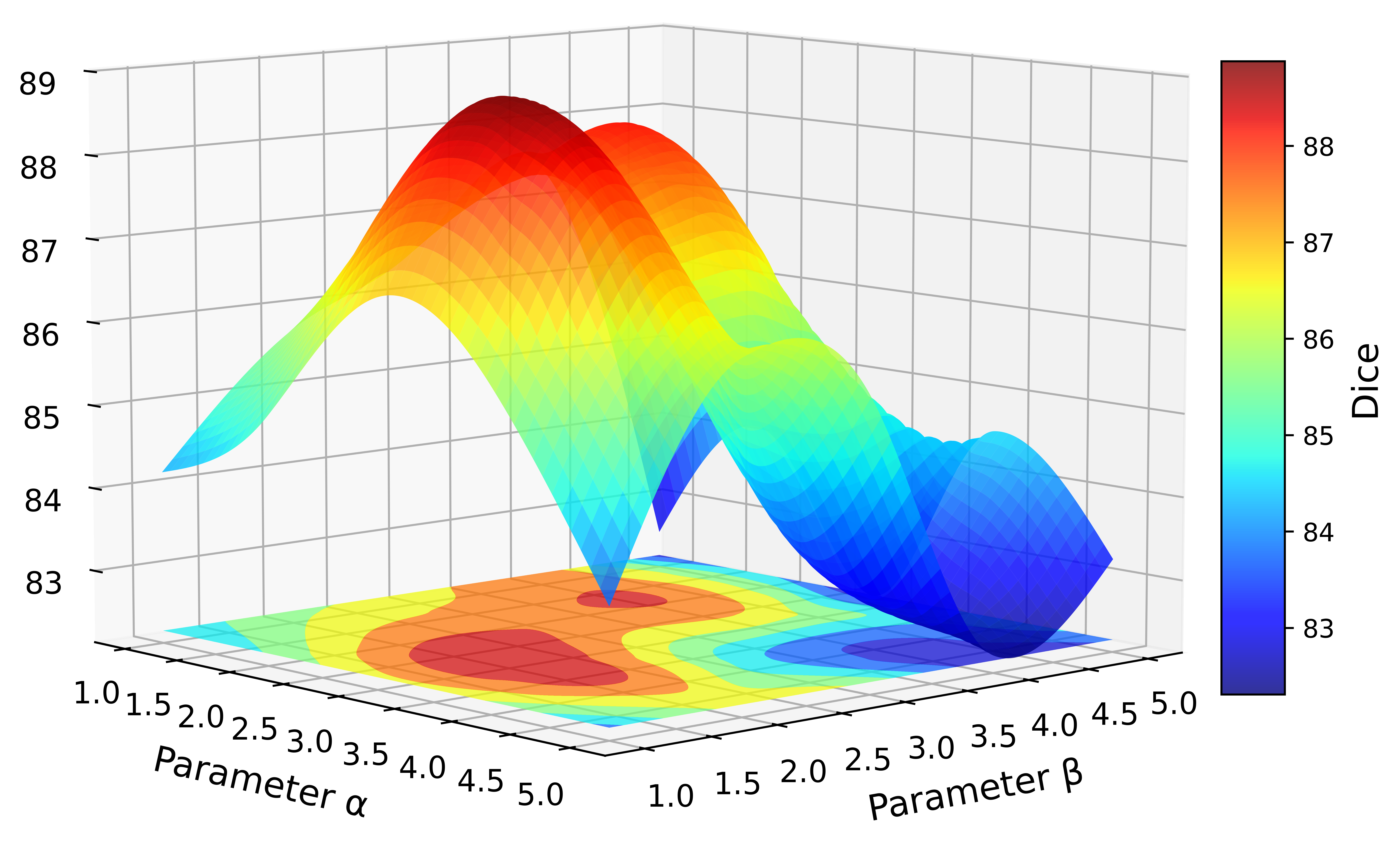}
        \vspace{-2em}
	\caption{Visualization of the impact of different $\alpha$ and $\beta$ values on NIH pancreas segmentation dataset under 10\% HQ label data and 90\% LQ label data. Where C denotes Dice score.}\label{fig6}
\end{figure} 


\subsection{Parameter Sensitivity Analysis} \label{sec4.5}



To assess the robustness and generalizability of our method, we conducted extensive sensitivity analysis on key hyperparameters involved in our framework. Specifically, we focus on two aspects: the sample selection ratio $k$, and the loss weighting parameters $\alpha$ and $\beta$. 

\noindent \textbf{Impact of $k$ on Training Dynamics.}
The parameter $k$ controls the proportion of low-quality (LQ) labeled samples used in each mini-batch. As shown in Fig.~\ref{fig5}, both overly small and large $k$ values degrade performance. Small $k$ leads to underutilization of useful LQ data, while large $k$ increases the risk of incorporating unreliable refined labels, introducing noise during optimization. The best performance is achieved at $k = 0.5$, which effectively balances label quality and data diversity. This demonstrates the importance of selective sample inclusion and confirms the robustness of our label selection strategy.

\noindent \textbf{Influence of $\alpha$ and $\beta$.}
We also analyzed the effects of $\alpha$ and $\beta$, which weight the label selection loss $\mathcal{L}{ls}$ and noisy label loss $\mathcal{L}{n}$, respectively. A grid search over $\alpha, \beta \in [1,5]$ on the NIH dataset under 10\% HQ / 90\% LQ conditions yielded the results shown in Fig.~\ref{fig6}. The surface plot reveals nonlinear sensitivity, where both overly low and high values reduce Dice performance, confirming the need for balanced loss weighting. The optimal region centers around $\alpha=3$, $\beta=2$, with stable performance in nearby ranges (e.g., $\alpha \in [2.5, 3.5]$). In contrast, high values ($>4$) lead to overfitting due to excessive reliance on HQ labels, while low values ($<1.5$) weaken the benefits of adaptive supervision. Overall, these results validate the effectiveness and stability of our loss design. Future work may explore adaptive or learnable weighting schemes to further enhance flexibility and performance across varying noise conditions.

\section{Conclusion}
This paper introduces an Adaptive Label Correction (ALC) framework designed for robust medical image segmentation. The framework employs a dynamic label refinement algorithm to adaptively weight multiple perturbation variants, improving the quality of noisy label data. Additionally, the overall confidence of each noisy labeled sample is evaluated, enabling the selective learning of high-confidence samples under low-quality label learning, while the remaining samples are processed through noisy label learning strategies. Comprehensive experiments on two public datasets demonstrate the effectiveness of each component in our proposed approach. Compared to the state-of-the-art methods, our framework demonstrates superior performance and robust competitiveness in managing noisy label data.

\clearpage

\bibliographystyle{IEEEtran}
\bibliography{ref}

\end{document}